\documentclass[letterpaper,10pt,conference]{ieeeconf}

\newif\ifshowappendix
\showappendixtrue

\usepackage{amsmath,amssymb}
\usepackage{booktabs}
\usepackage{capt-of}
\usepackage{graphicx}
\usepackage{multirow}
\usepackage{tabularray}
\usepackage{xcolor}
\usepackage{url}
\usepackage[hidelinks]{hyperref}

\title{\LARGE\bfseries
mmHRI: Towards Privacy-Preserving Human-Robot Interaction
with Millimeter-Wave Radar
}
\author{Junqiao Fan$^{1}$, Yuxuan Hu$^{2}$, Bofan Lyu$^{2}$, Yanshuo Lu$^{2}$, Pengfei Liu$^{2}$,\\
Jiarui Zhang$^{1}$, Fangqiang Ding$^{3}$, Lihua Xie$^{1}$, Gen Li$^{4,\dagger}$, Jianfei Yang$^{2,\dagger}$%
\thanks{$^{1}$School of Electrical and Electronic Engineering, Nanyang Technological University.
$^{2}$School of Mechanical and Aerospace Engineering, Nanyang Technological University.
$^{3}$The Hong Kong University of Science and Technology (Guangzhou).
$^{4}$School of AI and Robotics, Hunan University.
$^{\dagger}$Corresponding authors: \texttt{ligen@g.skku.edu}; \texttt{jianfei.yang@ntu.edu.sg}.}%
\thanks{Project website: {\urlstyle{same}\url{https://fanjunqiao.github.io/mmHRI-site/}}.}%
}

\IEEEoverridecommandlockouts

\makeatletter
\let\mmhri@originalmakecaption\@makecaption

\long\def\@makecaption#1#2{%
  \ifx\@captype\@IEEEtablestring
    \parbox{\linewidth}{%
      \normalfont\normalsize
      \noindent #1: #2\par
    }%
    \vskip\abovecaptionskip
  \else
    \mmhri@originalmakecaption{#1}{#2}%
  \fi
}
\makeatother

\begin{document}

\makeatletter
\let\mmhri@originalmakefntext\@makefntext
\long\def\@makefntext#1{\parindent\z@\noindent\hbox{\@makefnmark}#1}
\maketitle
\let\@makefntext\mmhri@originalmakefntext
\makeatother
\thispagestyle{empty}
\pagestyle{empty}

\begin{abstract}
Assistive robots increasingly operate in many human-centered environments and perform various human-robot interaction (HRI) tasks, such as object delivery. However, most existing HRI systems rely on RGB cameras that continuously observe humans to respond to non-verbal commands, such as hand gestures. This raises privacy concerns in privacy-critical environments, such as hospital wards or restaurants, where direct camera observation of humans is restricted. To develop privacy-preserving HRI, we leverage millimeter-wave (mmWave) radar, which can sense human motion through privacy barriers without identifiable imagery. We propose mmHRI, the first multi-modal robot manipulation framework that achieves mmWave radar-guided privacy-preserving HRI. mmHRI introduces two key designs to mitigate the sparsity and temporal inconsistency of radar data in cluttered robot manipulation environments. First, we propose a dual-stream architecture that jointly learns from unfiltered raw radar tensors and radar point clouds to estimate both human actions and 3D poses. To mitigate signal inconsistency, mmHRI further incorporates a memory-based state-space model (MSSM) that retains historical radar features to reduce abrupt changes in pose/action. These estimated human states are then converted into structured textual robot instructions, which control a vision-language-action (VLA) policy for closed-loop robot manipulation and human-aware reactions. Our evaluation covers human action recognition and closed-loop delivery and retrieval. In the privacy-preserving curtain setting, mmHRI achieves 85.09\% action-recognition accuracy, outperforming existing radar-based alternatives. Robot trials further demonstrate successful delivery and retrieval under visual occlusion, with stable task performance across unseen subjects, clutter configurations, and environments.

\end{abstract}

\suppressfloats[t]
\begin{figure}[t]
    \centering
    \includegraphics[width=\columnwidth]{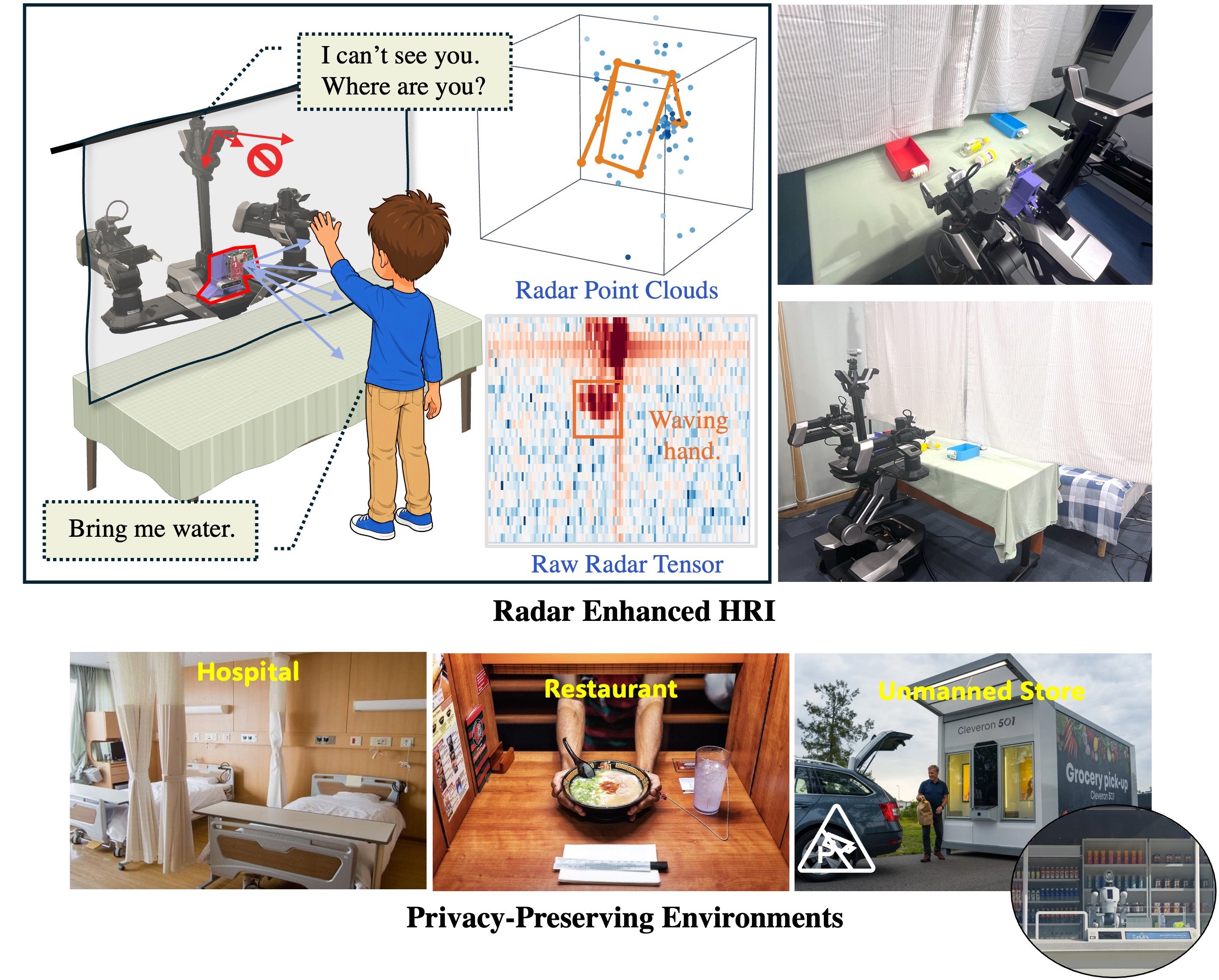}
    \vspace{-1.8em}
    \caption{Motivation of privacy-preserving mmHRI using mmWave radar for human sensing. Unlike conventional
    vision-based HRI that fails under occlusion, mmHRI senses human intent through privacy curtains to coordinate object
    delivery without directly observing humans. This supports
    privacy-sensitive applications such as hospitals, restaurants, and unmanned stores.}
    \vspace{-1em}
    \label{icra:fig:teaser}
\end{figure}

\section{Introduction}
\label{icra:sec:introduction}
Recent advances in perception, manipulation~\cite{chi2025diffusion}, and robot foundation
models~\cite{kim2024openvla,intelligence2025pi_} have accelerated the deployment of
service robots in hospitals, restaurants, retail stores, and other human-centered environments.
In these applications, effective human--robot interaction (HRI) requires robots to understand not
only object states but also human intent, including where and when users expect a response. Although verbal instructions
can specify intent, humans also convey much of their intent through non-verbal
cues~\cite{birdwhistell2010kinesics}, such as hand gestures. For example, during
object delivery, a user may gesture toward one side to indicate a preferred delivery location or
wave a hand to signal readiness.

Most existing HRI systems rely on RGB-D cameras to perceive non-verbal human interaction cues.
However, RGB-based systems require direct visual observation of the interacting person, leading to
two practical limitations. First, cameras capture faces, clothes, and body images that raise \textbf{privacy concerns}. Therefore, direct visual observation is often restricted in privacy-sensitive environments~\cite{baselizadeh2024prima,liang2025openrobocare,liu2025understanding}.
For example, privacy curtains in hospital wards intentionally shield patients from observation
during rest and treatment. Similarly, wooden partitions in some Japanese restaurants
(e.g., ICHIRAN) separate customers from staff and other diners to ensure dining privacy. These privacy barriers make it more challenging for robots to perceive user intent and respond promptly without compromising
privacy. Moreover, varying illumination conditions, such as darkness and strong
sunlight, may also degrade visual observations and reduce the reliability of human state perception~\cite{chen2022mmbody,ho2024rt}.

Millimeter-wave (mmWave) radar has emerged as a promising and affordable complement to existing
line-of-sight (LoS) camera-based systems. Sensitive to moving targets, mmWave radar captures
human motion by transmitting and receiving radio-frequency (RF) signals without revealing
identifiable visual information (e.g., facial appearance or clothing)~\cite{zhang2023survey}.
Its signals can penetrate certain visually opaque materials (e.g., curtains or wooden
partitions), and is more robust to illumination changes~\cite{xue2021mmmesh}.
These properties have motivated various radar-based human sensing applications~\cite{singh2019radhar,zhao2019mid,yang2023mm}.
However, most existing radar systems are developed for clean, uncluttered monitoring 
environments. Using radar for HRI and robot manipulation is more challenging.
First, commercial radar has lower resolution and produces
\textbf{sparse radar point clouds (RPC)} around moving human body parts~\cite{fan2024diffusion}.
Detected body parts may occasionally disappear due to specular reflections~\cite{ding2024milliflow},
causing temporal inconsistencies in RPCs.
Second, robot-manipulation environments introduce additional
\textbf{sensing interference (clutter)} compared with conventional monitoring environments.
Furniture (e.g., tables), delivered objects, privacy curtains, and moving robot arms can generate
multipath reflections and false ghost targets~\cite{sun20213drimr}, further increasing temporal
inconsistency in human perception. 

To achieve privacy-preserving HRI, we propose \textbf{mmHRI}, the first learning-based radar-vision multimodal 
HRI framework that integrates mmWave radar as an additional human-perception
modality. mmHRI is built on a
vision-language-action (VLA) backbone and can be implemented with any off-the-shelf alternatives. Its RGB cameras observe only the robot working area for
fine-grained object manipulation. Meanwhile, mmWave radar captures human non-verbal intent
(e.g., gestures and poses) to determine
where and when the robot should react. To compensate for RPC sparsity, 
we first propose dual-stream radar-based human perception (DRP), exploring the raw radar tensor and its fusion with RPC. Specifically, the raw
Range--Doppler--Time radar heatmap preserves more unfiltered temporal
micro-velocity information for human action/gesture estimation, while the filtered 3D RPC retains 3D geometry information for human pose estimation. 
To further mitigate temporal inconsistency
in radar signals, we introduce a memory state-space model (MSSM) to track
human states over time and memorize historical radar patterns. This prevents occasional curtain
motion or robot-arm motion from being misinterpreted as human commands. Finally, human-aware
text reasoning (HTR) converts the radar-estimated human states, including gestures and human poses, into text-based robot reaction descriptions. These structured text commands
are fed into a VLA policy for closed-loop manipulation that adapts to
real-time human intent.

We evaluate mmHRI using both real-world robot trials and our collected human perception dataset
for privacy-preserving HRI. mmHRI achieves higher accuracy and robustness than existing radar-based
systems. The robot trials further demonstrate that mmHRI is more robust under occlusion than existing
RGB-based systems. Extensive experiments also show robustness across unseen subjects, clutter,
and environments. The main contributions of this work are as follows:

\begin{enumerate}
\item We present \textbf{mmHRI}, the first multimodal HRI framework that integrates mmWave radar human perception into a robot manipulation foundation model, achieving closed-loop privacy-preserving object delivery.

\item We introduce a dual-stream radar perception (DRP) architecture and a memory state-space model (MSSM) for radar-based human perception, tailored for cluttered robot manipulation environments.

\item We conduct systematic experiments using our collected HRI-oriented human perception dataset and real-world robot
trials to evaluate human perception performance and its downstream effects on robot decision-making.

\end{enumerate}

\begin{figure*}[t]
    \centering
    \includegraphics[width=.9\textwidth]{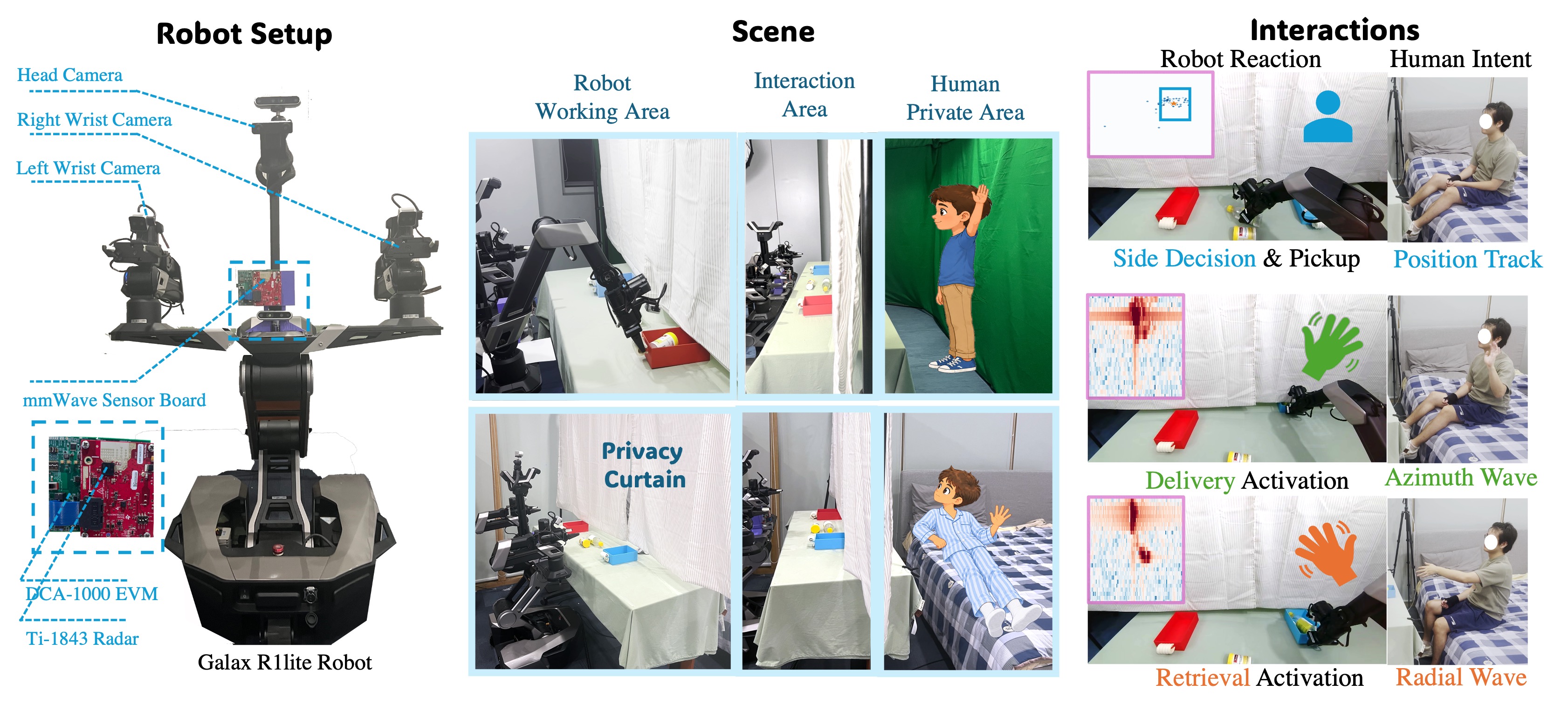}
    \vspace{-1em}
    \caption{Left: Physical robot system setup and the mmWave sensing unit. Middle: Two privacy-preserving scenes, i.e., hospital ward and office. Right: Three gesture/pose based human-robot-interaction decision-making tasks. We show different radar signals that the robot may refer to for different reactions.}
    \label{icra:fig:scene-setup}
\end{figure*}

\section{Related Work}
\label{icra:sec:related_work}

\subsection{Radar-Based Human Perception}
Human action recognition (HAR), human pose estimation (HPE), and other human-perception technologies
are important computer-vision tasks for various applications, such as
virtual reality~\cite{keller2023skin}, surveillance~\cite{kim2019skeleton}, and
HRI~\cite{reily2018skeleton}. Most existing methods rely on RGB-D camera systems~\cite{maji2022yolo,shi2019two}, which remain limited by visual
occlusion, changing illumination~\cite{chen2022mmbody}, and privacy concerns~\cite{an2021mars,liu2025understanding}.
Radar has therefore been adopted as a complementary or alternative modality for human sensing. Early works
focus on coarse, clustering-based human tracking~\cite{zhao2019mid, gu2019mmsense} from sparse radar point clouds (RPCs).
Subsequent works~\cite{singh2019radhar,hu2026waveman} explore RPC-based action and gesture recognition using deep-learning algorithms.
Recent methods further explore higher-fidelity pose estimation using RPCs~\cite{xue2021mmmesh,chen2022mmbody,yang2023mm} or raw radar tensors~\cite{ho2024rt,xue2023towards}.
Nevertheless, these methods typically assume clean, uncluttered sensing environments, whereas HRI and robot-manipulation scenarios
are usually severely cluttered. Several works have integrated mmWave radar into robotic tasks: WaveMan~\cite{hu2026waveman}
provides RPC-based recognition of four gestures for teleoperating robots in an open room. OmniVLA~\cite{guo2025omnivla} incorporates radar to locate non-line-of-sight objects inside boxes. It applies a simple masked overlay of radar measurements
and RGB observations to identify which object to grasp.
Yet, these methods remain restricted to static objects or relatively clean and uncluttered environments.

\subsection{Learning-Based Human--Robot Interaction}
Traditional human--robot interaction methods~\cite{strabala2013toward,phan2025placing,choi2009hand} typically require RGB-based pose estimation and subsequent
action recognition to estimate the human state. They typically rely on predefined robot responses for tasks such as human following~\cite{fujii2014gesture} and object delivery~\cite{phan2025placing,choi2009hand,mon2025embodied}.
Recently, learning-based robot policies have achieved better generalization in object manipulation. Diffusion
Policy learns visuomotor control directly from visual observations~\cite{chi2025diffusion}, while
OpenVLA~\cite{kim2024openvla} and $\pi_{0.5}$~\cite{intelligence2025pi_} connect visual observations and language instructions to robots.
However, these policies generally
lack human understanding and rely on detailed language descriptions, which is inefficient for HRI. HABIT~\cite{song2026habit}, Gaze2Act~\cite{zuo2026gaze2act} and more~\cite{mon2025embodied,liu2026give,li2026gazevla} incorporate human non-verbal information into VLA policies for
HRI, but require direct RGB observation of the human or a clear view for
gesture and gaze recognition. Consequently, these methods fail when humans are outside the
visual line of sight under privacy-sensitive scenarios. Several methods~\cite{wang2018controlling} design wearable sensing systems to support
non-line-of-sight interaction but require strong user compliance with wearing the devices.
Therefore, incorporating radar into HRI is promising to
address remote privacy-preserving human perception under visual occlusion, but its real-world deployment for HRI remains underexplored.


\section{System Overview}
\label{icra:sec:system overview}

\noindent\textbf{Robot Setup.}
As presented in Figure~\ref{icra:fig:scene-setup}, our platform is built on a R1 Lite dual-arm mobile robot. The robot is equipped with a head camera that focuses on the working table. 
Two wrist cameras are mounted on the robot's left and right arms to support fine-grained object manipulation. 
Human perception is achieved by a mmWave radar sensing board mounted on the robot's chest, which can operate under 
darkness or through occlusions.
It consists of a TI IWR1843 mmWave radar (3TX 4RX) and a DCA1000 data-capture board for 10 Hz raw data acquisition. 
The sensing board and robot transmit synchronized radar and camera
observations to a Linux policy server, which performs perception, reasoning, and policy inference. It then generates action chunks and sends them back to the robot for execution.

\noindent\textbf{Scene Setup.}
Our scenes are set up for privacy-preserving human--robot interaction (HRI) in two different environments, simulating hospital and office settings. 
As illustrated in Figure~\ref{icra:fig:scene-setup}, each scene comprises three
areas: (1) The robot working area is observed by the robot's RGB cameras to support object
manipulation. The fields of view of all onboard cameras are restricted
to this area; (2) The interaction area is accessible by both the robot and the human, including a table 
for conducting object delivery and a curtain for privacy protection; and (3)
The human private area is occupied only by the human and cannot be directly monitored by RGB cameras. 
This physical setup inevitably contains several sources of radar interference, including the
table, robot hardware, curtain, and nearby objects. 

\begin{figure*}[t]
    \centering
    \includegraphics[width=0.9\textwidth]{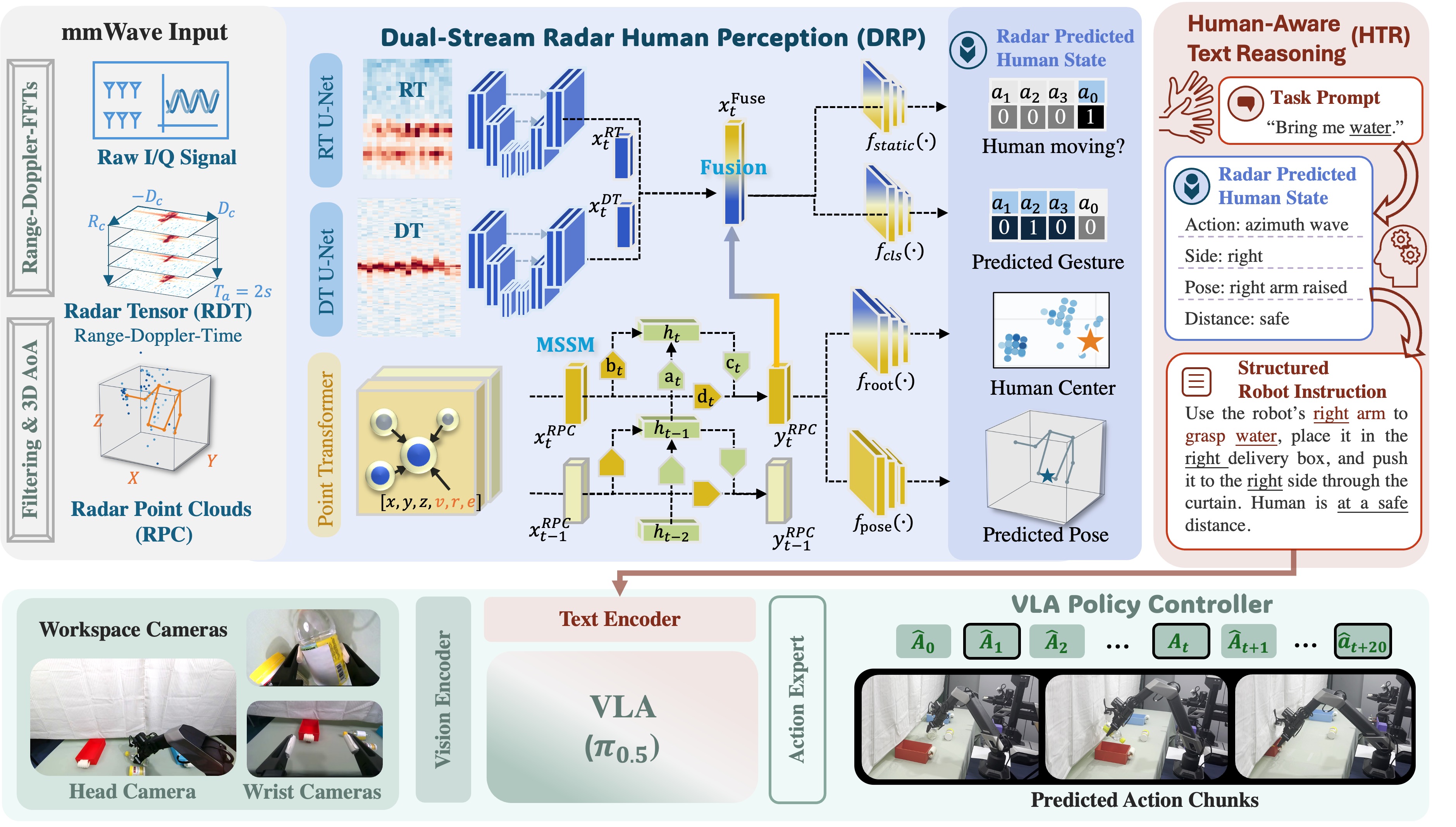}
    \caption{Overview of mmHRI. Radar measurements are first preprocessed into RDT
    tensors and RPC. The dual-stream radar human perception (DRP) then extracts motion and geometry patterns from both modalities, which 
    are fused to jointly predict action and 3D poses. The human-aware text reasoning (HTR) then converts the estimated human states into structured robot instructions, controlling the downstream VLA policy for different actions.}
    \label{icra:fig:system_task}
\end{figure*}

\noindent\textbf{Problem Formulation.}
We study a classic human--robot interaction task: tabletop/bedside object delivery.
The robot is tasked to deliver human-specified objects to their reachable region, understanding gesture intent, such as where and when the delivery 
should take place.
First, the robot identifies the requested object and uses the user's location and gestures to select the delivery side (Where).
Second, after an azimuth-wave greeting gesture (When), it grasps the object, places it in a delivery box, and pushes the box within the user's reach. 
Finally, it waits and retrieves the delivery box until it recognizes a radial wave (When).
Throughout the task, it continuously monitors the user's position and pauses actions whenever the user is near the robot to avoid collisions.

We formulate this HRI task as a learning problem. Given radar observations $O_{\mathrm{radar}}$, RGB observations $O_{\mathrm{rgb}}$,
and a simple spoken task description $T_{\mathrm{task}}$, the mmHRI policy $\pi$
outputs a 20-step action chunk $A$ for the two robot arms:
$
A=\pi(O_{\mathrm{radar}},O_{\mathrm{rgb}},T_{\mathrm{task}}),
A\in\mathbb{R}^{T_{\mathrm{chunk}}\times 2N_{\mathrm{arm}}},
$
where $T_{\mathrm{chunk}}=20$ and $N_{\mathrm{arm}}=7$ in our physical robot setup.
The task description specifies the requested object, e.g., \textit{``Bring me water.''}
The radar observations $O_{\mathrm{radar}}$ are raw I/Q signals from the mmWave sensor
board for real-time human-state perception.
The RGB observations $O_{\mathrm{rgb}}$ are multi-view images from robot-mounted
cameras for object manipulation.

\section{Method}
\label{icra:sec:method}

As presented in Figure~\ref{icra:fig:system_task}, we first preprocess the input raw radar signals into a radar tensor and a radar point cloud
(Sec.~\ref{icra:subsec:radar-preprocessing}).
We then use a dual-stream radar perception network to estimate human actions, 3D pose, and root location
(Sec.~\ref{icra:subsec:human-state-estimation}).
Next, we design a human-aware text reasoning (HTR) module that converts the estimated human state
and task description into a structured text prompt (Sec.~\ref{icra:subsec:human-state-vla}).
Finally, this prompt conditions a single VLA policy to generate different robot reactions
(Sec.~\ref{icra:subsec:human-state-vla}).

\subsection{Radar Signal Preprocessing}
\label{icra:subsec:radar-preprocessing}
We process the raw complex ADC signals into two representations for human-state estimation:
a range--Doppler--time (RDT) tensor and a radar point cloud (RPC).
To construct the raw radar RDT tensor, we first generate range--Doppler (RD) maps by applying a range FFT and a Doppler FFT
to the ADC signals. We apply mean pooling across all virtual antennas to reduce computation.
Following mmMesh~\cite{xue2021mmmesh}, we perform chirp-wise mean subtraction on the range maps to remove background clutter.
The RD map produces a heatmap that is sensitive to micro-Doppler motion responses,
providing information about how far the human is from the robot and how the human moves.
Finally, we stack $T_a = 20$ RD maps spanning 2\,s to construct the RDT tensor.
To obtain the RPC, we use top-K ($K=128$) energy selection to detect targets with the most salient velocity responses.
Then, we calculate the angle of arrival (AoA) to obtain the azimuth and elevation angles of the detected targets,
generating the 3D RPC.
We observe that human motion produces stronger reflections in the RD map, whereas multipath reflections generally lose energy.
Therefore, we apply an additional energy threshold to further suppress multipath points.
Finally, following previous works~\cite{yang2023mm}, we interpolate RPCs from four historical frames into one frame to reduce sparsity.



\subsection{Dual-Stream Radar-Based Human State Estimation}
\label{icra:subsec:human-state-estimation}
Dual-stream radar-based human perception uses both the unfiltered RDT tensor
$O^{\mathrm{RDT}}$ and the filtered RPC $O^{\mathrm{RPC}}$
for human-state estimation. It designs a multi-task framework, simultaneously performing
action classification, root tracking, and pose estimation. 
Specifically, we use the RDT tensor for high-level, spatially agnostic action recognition. 
Meanwhile the filtered RPC supports pose estimation and root tracking because it preserves 3D spatial
information.

\begin{figure}[!t]
    \centering
    \includegraphics[width=\columnwidth]{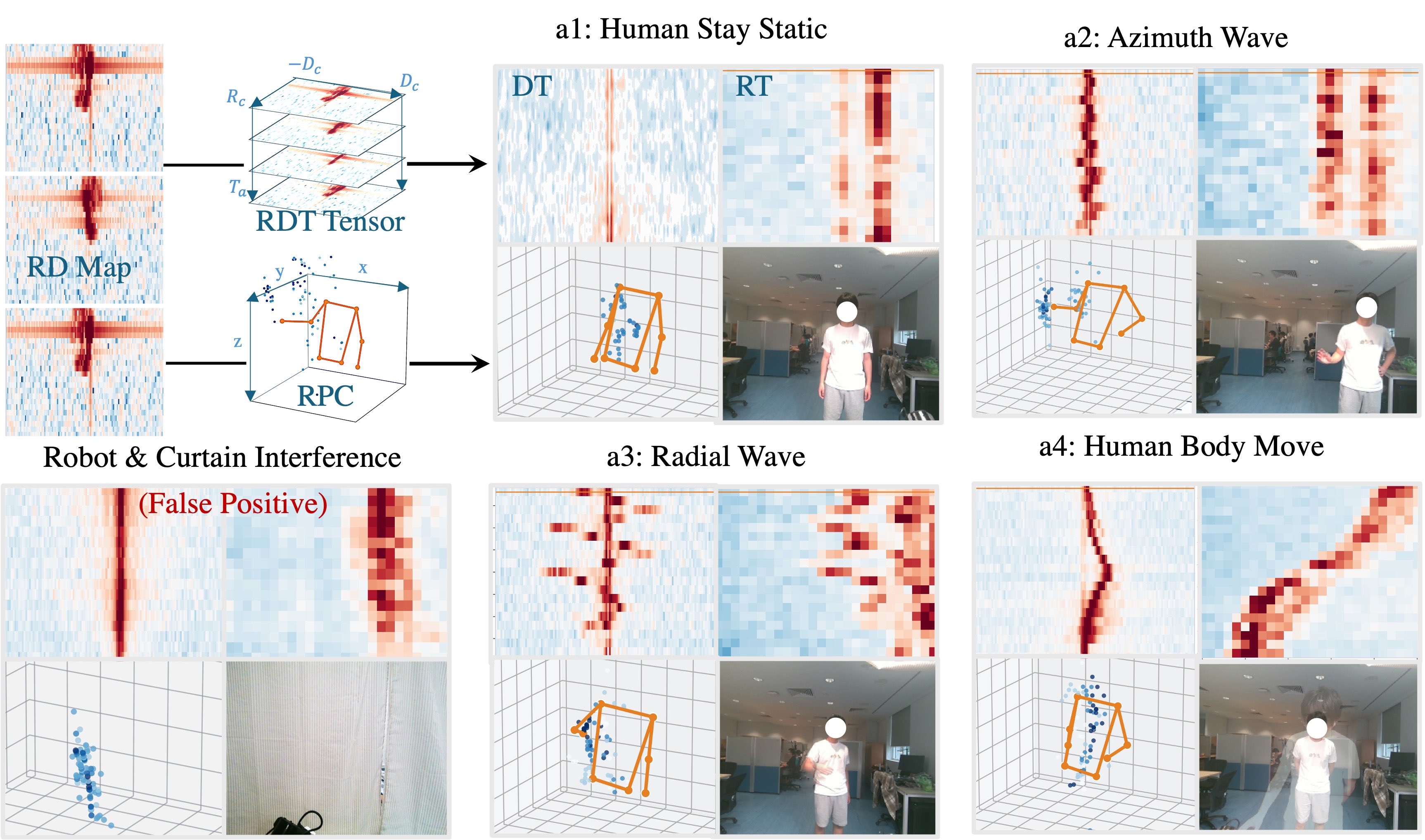}
    \caption{Visualization of the complementary radar modalities used by the dual-stream
    model. The DT and RT maps preserve motion-sensitive Doppler patterns, whereas the RPC retains
    3D spatial structure for pose and root estimation. Synchronized RGB images are shown for
    visual reference. }
    \vspace{-1em}
    \label{icra:fig:radar-modalities}
\end{figure}

\noindent\textbf{RDT Feature Extraction.}
The RDT stream receives $O^{\mathrm{RDT}}$ as inputs. The RDT tensor is higher-dimensional 
and generally retains more unfiltered motion-related information than the standard RPC.
However, directly applying 3D convolution over the entire tensor is
computationally inefficient. Therefore, we reshape the tensor and apply two 2D convolution U-Nets along two viewing directions, 
Range-Time (RT) and Doppler-Time (DT). 
The additional dimensions are treated as feature channels. As illustrated in
Figure~\ref{icra:fig:radar-modalities}, the RT and DT representations capture subtle micro-motion responses
for different hand-waving gesture actions. Specifically, we use efficient three-layer RT and DT U-Nets for feature
extraction, producing $x_t^{\mathrm{RT}}$ and $x_t^{\mathrm{DT}}$ at time $t$.

\noindent\textbf{Memory State-Space-Model (MSSM).}
At time $t$, the RPC stream receives $O_t^{\mathrm{RPC}}$ and applies a Point
Transformer~\cite{zhao2021point} to perform point-wise self-attention and extract the RPC feature $x_t^{\mathrm{RPC}}$.
To mitigate temporal inconsistency in radar signals, we design a state-space model (ssm) 
to enhance the RPC feature with historical memory.
The memory state $h_{t}$ stores both current-frame $x_t^{\mathrm{RPC}}$ information and historical information from previous $h_{t-1}$.
As shown in Figure~\ref{icra:fig:system_task}, the state is updated using four
coefficients that depend on the current RPC feature $x_t^{\mathrm{RPC}}$:
\begin{equation}
\begin{aligned}
a_t &= g_a(x_t^{\mathrm{RPC}}), & b_t &= g_b(x_t^{\mathrm{RPC}}), \\
c_t &= g_c(x_t^{\mathrm{RPC}}), & d_t &= g_d(x_t^{\mathrm{RPC}}),
\end{aligned}
\label{icra:eq:mssm-coefficients}
\end{equation}
where $g_a$, $g_b$, $g_c$, and $g_d$ are learnable one-layer MLPs.
The memory state evolves as:
\begin{equation}
h_t = a_t h_{t-1} + b_t x_t^{\mathrm{RPC}}.
\label{icra:eq:mssm-state}
\end{equation}
The output of MSSM $y_t^{\mathrm{RPC}}$ is then calculated as:
\begin{equation}
y_t^{\mathrm{RPC}} = c_t h_t + d_t x_t^{\mathrm{RPC}}.
\label{icra:eq:mssm-output}
\end{equation}
This recurrence design carries information across historical frames while retaining current spatial features, preventing
subtle changes caused by occasional signal inconsistency.

\noindent\textbf{Dual-Stream Feature Fusion.}
We concatenate the RDT features with the MSSM output along the feature dimension:
\begin{equation}
x_t^{\mathrm{Fuse}} = [x_t^{\mathrm{RT}};x_t^{\mathrm{DT}};y_t^{\mathrm{RPC}}].
\label{icra:eq:radar-fusion}
\end{equation}
This feature combines both unfiltered motion information from RDT and spatial information from RPC.
As shown in Figure~\ref{icra:fig:system_task}, the fused feature is then utilized by four MLP heads to estimate four different human states: First, a static head $f_{\mathrm{static}}$ predicts whether a non-static human action is present. Meanwhile, an action classification head $f_{\mathrm{cls}}$ predicts the logits for three non-static
actions: azimuth wave, radial wave, and body move:
\begin{equation}
\begin{aligned}
\hat{p}_{\mathrm{static},t}
&= \operatorname{sigmoid}\!\left(f_{\mathrm{static}}(x_t^{\mathrm{Fuse}})\right),\\
\hat{p}_{\mathrm{act},t}
&= \operatorname{softmax}\!\left(f_{\mathrm{cls}}(x_t^{\mathrm{Fuse}})\right).
\end{aligned}
\label{icra:eq:static-head}
\end{equation}
To obtain the 3D human position, the root tracking head $f_{\mathrm{root}}$ first predicts the coordinates of the human root:
\begin{equation}
\hat{\boldsymbol{\tau}}_t = f_{\mathrm{root}}(x_t^{\mathrm{Fuse}})\in\mathbb{R}^{3}.
\label{icra:eq:root-head}
\end{equation} 
Finally, the pose estimation head $f_{\mathrm{pose}}$ predicts the root-normalized 3D positions of
eight upper-body joints:
\begin{equation}
\hat{P}_t = f_{\mathrm{pose}}(y_t^{\mathrm{RPC}})\in\mathbb{R}^{8\times3}.
\label{icra:eq:pose-head}
\end{equation}
Together, $\hat{p}_{\mathrm{static},t},\hat{p}_{\mathrm{act},t},\hat{\boldsymbol{\tau}}_t,\hat{P}_t$ form the structured human states and are used by the downstream VLA policy.



\subsection{Human-State-Conditioned VLA Policy}
\label{icra:subsec:human-state-vla}

\noindent\textbf{Human-Aware Text Reasoning (HTR).}
As illustrated in Figure~\ref{icra:fig:system_task}, we convert the task text and radar-derived
human state into a structured robot instruction. The instruction template contains five fields:
the requested object from $T_{\mathrm{task}}$, and the action, receiving side, pose, and safety
distance from $S_{\mathrm{human}}=\{\hat{p}_{\mathrm{static},t},\hat{p}_{\mathrm{act},t},\hat{\boldsymbol{\tau}}_t,\hat{P}_t\}$.
The action class decoded from action predictions determines the current
interaction stage (e.g., delivery or retrieval); the root and hand locations determine the receiving side (e.g., left or right); 
and the human distance indicates whether execution is safe. These fields
are instantiated as an imperative instruction specifying which arm to use, which object to grasp,
which delivery box to use, and whether the robot may act. For example, the
task text ``Bring water'' and the human-state descriptions trigger the structured instruction $T_{\mathrm{robot}}$ 
shown in Figure~\ref{icra:fig:system_task}.
This structured robot prompt
$T_{\mathrm{robot}}$ directly controls downstream VLA policy for different robot action execution.

\begingroup
\definecolor{xls000000}{HTML}{000000}
\definecolor{xlsF2F2F2}{HTML}{F2F2F2}
\definecolor{xlsE8E8E8}{HTML}{E8E8E8}

\begin{table*}[t]
    \centering
    \caption{\textbf{Performance on the radar perception dataset.}
    HPE is evaluated under clear visibility, and HAR is evaluated under clear visibility, privacy occlusion, and cross-environment occlusion.}
    \label{icra:tab:simulated-performance}
\resizebox{.85\textwidth}{!}{\begin{tblr}{
  colspec={l *{14}{c}},
  cells={valign=m},
  colsep=4pt,
  row{4}={bg=xlsF2F2F2},
  row{6}={bg=xlsF2F2F2},
  row{10}={bg=xlsE8E8E8},
  cell{1}{1}={r=3}{l,m},
  cell{1}{2}={r=2,c=2}{c,m},
  cell{1}{4}={c=4}{c,m},
  cell{1}{8}={c=4}{c,m},
  cell{1}{12}={c=4}{c,m},
  cell{2}{4}={c=4}{c,m},
  cell{2}{8}={c=4}{c,m},
  cell{2}{12}={c=4}{c,m},
  cell{4,6,10}{1}={c=15}{c,m},
  hline{1}={1-15}{fg=xls000000},
  hline{4}={1-15}{fg=xls000000},
  hline{5}={1-15}{fg=xls000000},
  hline{6}={1-15}{fg=xls000000},
  hline{7}={1-15}{fg=xls000000},
  hline{10}={1-15}{fg=xls000000},
  hline{11}={1-15}{fg=xls000000},
  hline{Z}={1-15}{fg=xls000000},
  vline{2}={1-3,5,7-9,11-14}{fg=xls000000},
  vline{4}={1-3,5,7-9,11-14}{fg=xls000000},
  vline{8}={1-3,5,7-9,11-14}{fg=xls000000},
  vline{12}={1-3,5,7-9,11-14}{fg=xls000000},
}
  Methods & HPE &  & Clear Visibility &  &  &  & Occlusion (Privacy) &  &  &  & Occlusion (Cross-Env) &  &  &  \\
   &  &  & HAR &  &  &  & HAR &  &  &  & HAR &  &  &  \\
   & {MPJPE\\(cm)} & {TE\\(cm)} & {ACC\\(\%)} & {FPR\\(\%)} & {Static-F1\\(\%)} & {HPE\\Jitter} & {ACC\\(\%)} & {FPR\\(\%)} & {Static-F1\\(\%)} & {HPE\\Jitter} & {ACC\\(\%)} & {FPR\\(\%)} & {Static-F1\\(\%)} & {HPE\\Jitter} \\
  RGB &  &  &  &  &  &  &  &  &  &  &  &  &  &  \\
  RGB (YOLOv.) + AGCN & . & . & 96.02 & 4.55 & 97.06 & . & . & . & . & . & . & . & . & . \\
  mmWave Radar &  &  &  &  &  &  &  &  &  &  &  &  &  &  \\
  PointTrans. (RPC) + AGCN & 7.40 & 6.26 & 86.18 & 16.18 & 89.58 & 3.52 & 72.49 & 23.11 & 88.36 & 3.91 & 31.06 & 40.84 & 67.91 & 6.08 \\
  RadHAR (RPC) & . & . & 87.06 & 11.95 & 77.11 & . & 62.18 & 26.09 & 82.50 & . & 27.61 & 33.10 & 61.81 & . \\
  Waveman (RT) & . & . & 91.65 & 4.47 & 91.33 & . & 79.65 & 12.67 & 86.95 & . & 62.31 & 0.00 & 82.28 & . \\
  Ours &  &  &  &  &  &  &  &  &  &  &  &  &  &  \\
  mmHRI (Fusion) & 7.54 & 6.01 & 92.56 & 4.55 & 95.14 & 3.54 & 81.45 & \textbf{5.00} & 88.69 & 3.45 & 68.94 & 20.42 & 85.83 & 5.15 \\
  mmHRI (Fusion + MSSM) & \textbf{4.25} & \textbf{3.12} & \textbf{96.51} & \textbf{1.38} & \textbf{97.25} & \textbf{2.83} & \textbf{85.09} & 9.90 & \textbf{90.17} & \textbf{3.25} & \textbf{74.91} & \textbf{11.27} & \textbf{86.09} & \textbf{4.42} \\
\end{tblr}}
\end{table*}

\begin{table*}[t]
    \centering
    \caption{\textbf{Performance on real-world robot trials.}. PA and Success report the perception-only and end-to-end stage success rates. MA reports the success rate over 30 manipulation-only trials with ground-truth human states.}
    \label{icra:tab:robot-trials}
\resizebox{.85\textwidth}{!}{%
\begin{tblr}{
  colspec={*{13}{c}},
  cells={valign=m},
  row{3}={bg=xlsE8E8E8},
  row{6}={bg=xlsE8E8E8},
  row{9}={bg=xlsE8E8E8},
  cell{1}{1}={r=2}{c,m},
  cell{1}{2}={c=3}{c,m},
  cell{1}{5}={c=3}{c,m},
  cell{1}{8}={c=3}{c,m},
  cell{1}{11}={c=3}{c,m},
  cell{3}{1}={c=13}{c,m},
  cell{4}{3}={r=2}{c,m},
  cell{4}{6}={r=2}{c,m},
  cell{4}{9}={r=2}{c,m},
  cell{4}{12}={r=2}{c,m},
  cell{6}{1}={c=13}{c,m},
  cell{7}{3}={r=2}{c,m},
  cell{7}{6}={r=2}{c,m},
  cell{7}{9}={r=2}{c,m},
  cell{7}{12}={r=2}{c,m},
  cell{9}{1}={c=13}{c,m},
  cell{10}{3}={r=2}{c,m},
  cell{10}{6}={r=2}{c,m},
  cell{10}{9}={r=2}{c,m},
  cell{10}{12}={r=2}{c,m},
  hline{1}={1-13}{fg=xls000000},
  hline{3}={1-13}{fg=xls000000},
  hline{4}={1-13}{fg=xls000000},
  hline{6}={1-13}{fg=xls000000},
  hline{7}={1-13}{fg=xls000000},
  hline{9}={1-13}{fg=xls000000},
  hline{10}={1-13}{fg=xls000000},
  hline{12}={1-13}{fg=xls000000},
  vline{2}={1-2,4-5,7-8,10-11}{fg=xls000000},
  vline{5}={1-2,4-5,7-8,10-11}{fg=xls000000},
  vline{8}={1-2,4-5,7-8,10-11}{fg=xls000000},
  vline{11}={1-2,4-5,7-8,10-11}{fg=xls000000},
}
  Methods & Side Decision \& Grasping &  &  & Object Delivery  &  &  & Box Retrieval &  &  & Collision Avoidance &  &  \\
   & PA & MA & Success & PA & MA & Success & PA & MA & Success & PA & MA & Success \\
  Clear Visibility &  &  &  &  &  &  &  &  &  &  &  &  \\
  RGB + pi05 & 30/30 & 26/30 & 26/30 & 30/30 & 27/30 & 27/30 & 27/30 & 30/30 &27/30 & 30/30 & 30/30 & 30/30 \\
  mmWave+pi05 & 30/30 &  & 26/30 & 29/30 &  & 26/30 & 29/30 &  & 29/30 & 30/30 &  & 30/30 \\
  Occlusion &  &  &  &  &  &  &  &  &  &  &  &  \\
  RGB+pi05 & 0/30 & 26/30 & 0/30 & 0/30 & 27/30 & 0/30 & 0/30 & 30/30 & 0/30 & 0/30 & 30/30 & 0/30 \\
  mmWave + pi05 & 28/30 &  & 25/30 & 28/30 &  & 25/30 & 29/30 &  & 29/30 & 30/30 &  & 30/30 \\
  Occlusion (Cross-Environment) &  &  &  &  &  &  &  &  &  &  &  &  \\
  RGB+pi05 & 0/30 & 23/30 & 0/30 & 0/30 & 26/30 & 0/30 & 0/30 & 28/30 & 0/30 & 0/30 & 30/30 & 0/30 \\
  mmWave + pi05 & 28/30 &  & 21/30 & 25/30 &  & 22/30 & 26/30 &  & 25/30 & 25/30 &  & 25/30 \\
\end{tblr}%
}
\vspace{-1em}
\end{table*}
\endgroup

\noindent\textbf{VLA Control Policy.}
We adopt $\pi_{0.5}$~\cite{intelligence2025pi_} as the VLA backbone. At time $t$, the
visual observation contains one head-camera image and two wrist-camera images,
$O_t^{\mathrm{robot}}=\{O_{\mathrm{head},t},O_{\mathrm{lwrist},t},
O_{\mathrm{rwrist},t}\}$. The PaliGemma vision--language model
(VLM) encodes the images using a SigLIP-So400m visual encoder and
$T_{\mathrm{robot}}$ using a Gemma-2B text encoder. The resulting visual and language features are
jointly provided to a transformer-based flow-matching action head, which models the conditional
distribution of a 20-step action chunk for the two 7-DoF robot arms:
\begin{equation}
A_{t:t+20}\sim\pi_{0.5}\!\left(
\,\cdot\mid O_t^{\mathrm{robot}},T_{\mathrm{robot}}
\right),\qquad
A_{t:t+20}\in\mathbb{R}^{20\times14}.
\end{equation}
Here, $A_{t:t+20}$ denotes the action chunk over the interval $[t,t+20)$.
This interface enables a single policy to perform object selection, grasping, basket placement,
delivery, waiting, and retrieval according to the current human state.



\section{Experiments}
\label{icra:sec:experiments}

We design our experiments to answer three core questions. \textbf{Q1.} How does mmWave radar
benefit privacy-preserving and robust HRI in real-robot trials under clear vision and
visual occlusion?
\textbf{Q2.} How does the proposed mmHRI radar-perception model compare with existing radar-based methods? \textbf{Q3.} 
How robust is mmHRI under unseen human subjects, environments, and tabletop clutter setups?

\subsection{Experimental Details}

\noindent\textbf{Implementation Details.}
The VLA policy is initialized from the pretrained $\pi_{0.5}$ and fine-tuned with full parameters. We collect 160 real-world teleoperated demonstrations with
synchronized human-state descriptions (e.g., action, pose, root), structured robot instructions $T_{robot}$. The policy is optimized using AdamW for 1000 epochs with a learning rate of $1\times10^{-8}$ and a batch size of 64. Training takes 26 hours on four NVIDIA A800 GPUs. The DRP module adopts three-layer CNN U-Nets for the RDT tensor encoders, 
and the human state decoder are implemented with
two-layer MLPs. The entire DRP is pre-trained separately using AdamW for 10 epochs with a learning rate of $1\times10^{-4}$ and
a batch size of 64. We collect a radar perception dataset for DRP training and validation. The dataset includes 12K frames of a clear-visibility office scene (Clear Visibility); 2K testing frames of the same curtain-occluded office scene (Occlusion); and 2K testing frames of a curtain-occluded hospital ward scene (Occlusion Cross-Env). We use 80\% of Clear Visibility data for training and the remaining for testing. 


\noindent\textbf{Evaluation Metrics.}
For human action recognition (HAR), we report classification accuracy (ACC), false-positive
rate (FPR) and F1-score. The FPR measures the proportion of incorrect activation of static samples. The F1-score measures the performance over precision and recall. 
For human pose estimation (HPE), mean per-joint position error (MPJPE) measures
the average Euclidean error of root-normalized 3D joints, translation error (TE) measures the
Euclidean error of the human root, and jitter measures the frame-wise instability of
the predicted pose; all three are reported in centimeters. In real-robot trials, perception
accuracy (PA) measures the success rate of human intent detections during the end-to-end trials, including side decisions, delivery/retrieval
activation, and collision detections. Manipulation accuracy (MA) separately reports the VLA policy manipulation success rate over 30 manipulation-only trials using ground-truth human states. Finally, Success indicates the overall stage success rate during end-to-end trials. 

\subsection{Radar Perception Discussion}
As presented in Table~\ref{icra:tab:simulated-performance}, we compare mmHRI with existing radar-based perception methods. The RGB baseline first estimates human pose and then performs skeleton-based action recognition~\cite{shi2019two}. The RGB baseline achieves the best HAR performance under clear visibility, but cannot produce valid HAR or HPE predictions when humans are non-line-of-sight. In contrast, all mmWave radar-based methods perform HAR under both clear-visibility and occlusion conditions. Traditional RPC-based RADHAR~\cite{singh2019radhar} shows poor generalization, with ACC of only 62.18\% under occlusion and 27.61\% across environments. Replacing point-cloud input with a radar tensor and using the WaveMan~\cite{hu2026waveman} backbone improves ACC to 79.65\% under occlusion and 62.31\% across environments. This suggests that Doppler-rich motion patterns captured by RDT tensors are more informative and robust under cluttered HRI environments. However, radar tensors sacrifice spatial information and cannot support 3D spatial tasks such as HPE and tracking. mmHRI addresses this limitation through DRP fusion, which achieves joint 3D HPE and HAR. We observe that mmHRI achieves the best 85.09\% ACC under occlusion and 74.91\% ACC across environments, and reduces the FPR to below 11.27\%. The lower FPR reduces the risk that clutter or radar multipath interference is misclassified as an interaction gesture, which could trigger premature delivery/retrieval.
In addition, the MSSM module reduces HPE error and temporal jitter. By incorporating historical states and signal information, it mitigates abrupt changes and radar's temporal inconsistency, resulting in more stable spatial reasoning.


\begin{figure*}[t]
    \centering
    \includegraphics[width=.95\textwidth]{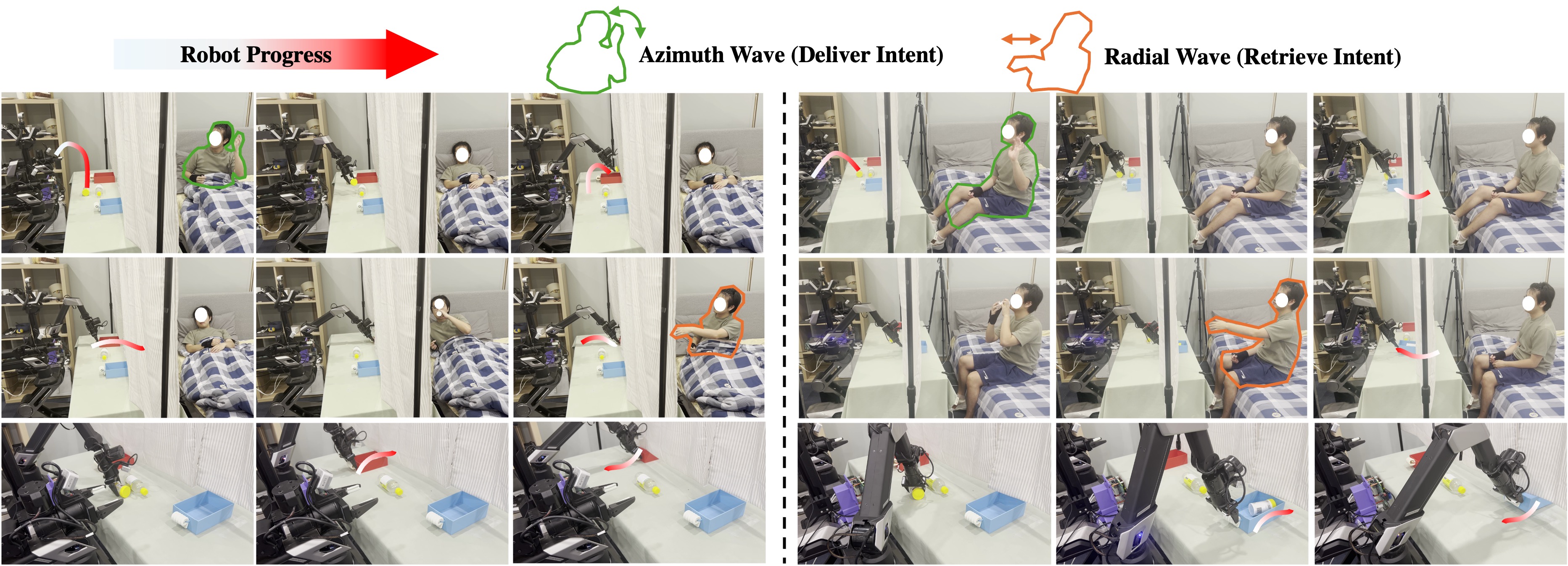}
    \caption{Qualitative examples of the human states and corresponding robot operations. The
    columns show an azimuth wave, waiting, approaching, and a radial wave; the camera views show
    object grasping, tray delivery, and tray retrieval during the interaction.}
    \vspace{-1em}
    \label{icra:fig:robot-interaction-sequence}
\end{figure*}

\subsection{Real-World Robot Trial Discussion}
As shown in Table~\ref{icra:tab:robot-trials}, our real-world robot trials evaluate both PA and MA. PA measures whether the robot recognizes human intents, and MA evaluates how many failures are attributed to the VLA policy. 
The RGB-based method performs well under clear visibility but fails under privacy-preserving occlusion scenes. In contrast, the mmWave radar-based method remains robust under these occluded conditions. Its PA performance also approaches RGB's under clear visibility. We observe that the radar method achieves slightly higher retrieval-activation PA. The radial waving can cause self-occlusion in RGB images, whereas radar captures hand motion patterns that are less affected by self-occlusion. 
For occlusion and cross-environment cases, the curtain motion may occasionally cause false activations and premature delivery/retrieval. The proposed DRP and MSSM modules aim to suppress these false positives and limit the performance degradation. For robot manipulation MA, occasional failures mainly occur when the relatively heavy water bottle loses balance and slips, or the delivery box collides with the table. The failure rate slightly increases in cross-environment trials due to changes in illumination. Since our mmHRI can be connected with any vision-language policies, the MA may improve with other alternatives. The radar-based method also achieves reliable collision detection, even when humans are blocked by privacy curtains, supporting safer HRI.

\begin{table}[!b]
    \centering
    \caption{\textbf{Generalization to unseen subjects.}}
    \label{icra:tab:subject-generalization}
    \footnotesize
    \setlength{\tabcolsep}{3.5pt}
    \renewcommand{\arraystretch}{1.12}
    \resizebox{\columnwidth}{!}{%
        \begin{tabular}{lccc}
            \toprule
            Subject & Delivery Activation & Retrieve Activation & Collision Avoidance \\
            \midrule
            Seen Subject   & 28/30 & 29/30 & 30/30 \\
            Unseen Subject 1 & 30/30 & 30/30 & 30/30 \\
            Unseen Subject 2 & 27/30 & 26/30 & 30/30 \\
            \bottomrule
        \end{tabular}%
    }
\end{table}

\begin{table}[!b]
    \centering
    \caption{\textbf{Generalization to unseen tabletop clutter.}}
    \label{icra:tab:clutter-generalization}
    \footnotesize
    \setlength{\tabcolsep}{3.5pt}
    \renewcommand{\arraystretch}{1.12}
    \resizebox{\columnwidth}{!}{%
        \begin{tabular}{lccc}
            \toprule
            Clutter & Delivery Activation & Retrieve Activation & Collision Avoidance \\
            \midrule
            No Additional Clutter         & 28/30 & 29/30 & 30/30 \\
            Metal Box    & 28/30 & 26/30 & 30/30 \\
            Tall Cup Box & 26/30 & 30/30 & 30/30 \\
            \bottomrule
        \end{tabular}%
    }
\end{table}

\begin{figure}[!hb]
    \centering
    \includegraphics[width=.9\columnwidth]{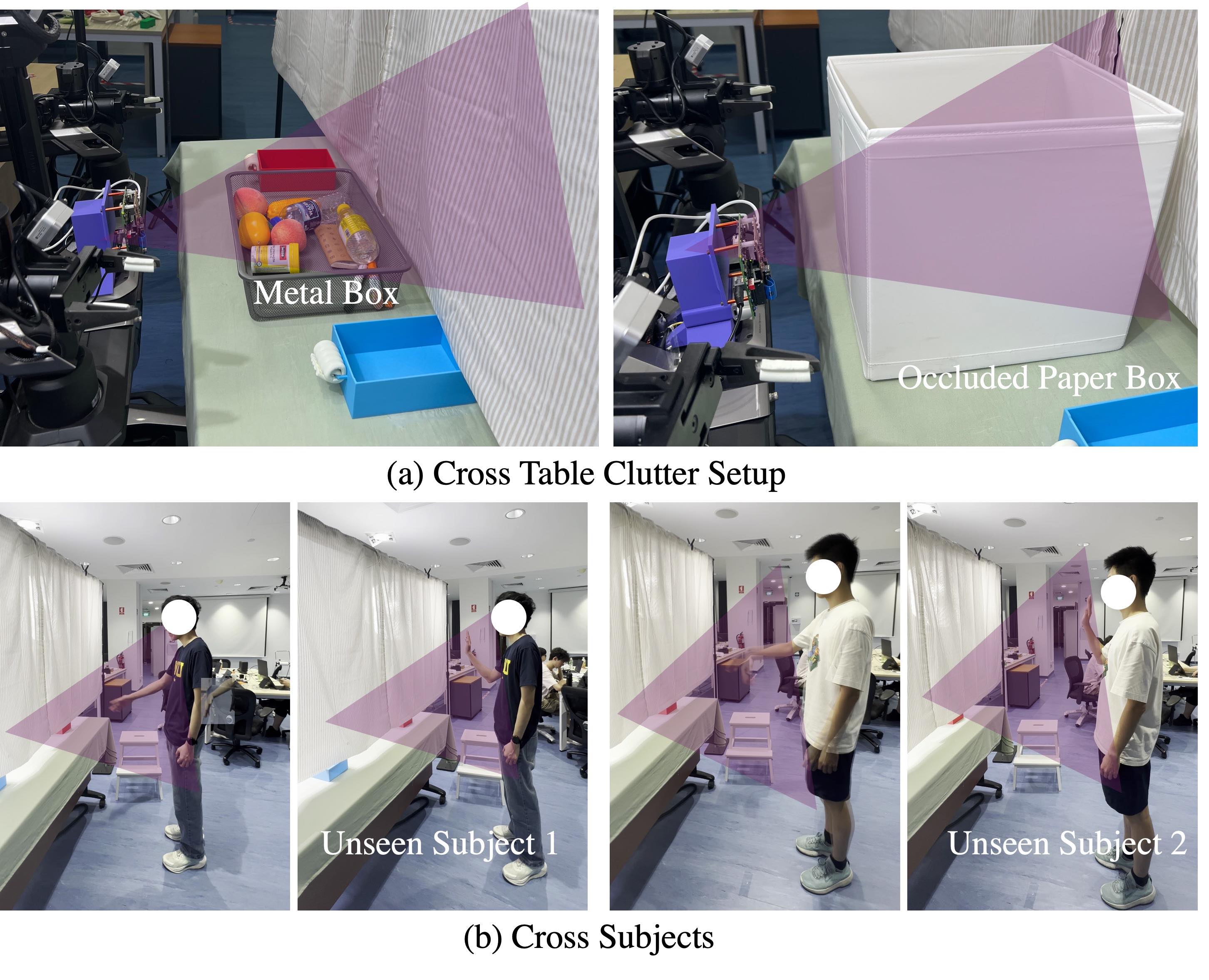}
    \vspace{-1.5em}
    \caption{Real-world robustness evaluation setups. (a) Cross-table-clutter settings with a
    metal box and an occluding paper box. (b) Cross-subject settings with two unseen subjects.}
    \label{icra:fig:robustness-setup}
\end{figure}

\subsection{Robustness and Generalization}
To evaluate generalization to unseen scenarios and assess deployment reliability, we conduct two
robustness tests for the perception module: cross-subject and cross-table-clutter
evaluation. As shown in
Figure~\ref{icra:fig:robustness-setup} and
Tables~\ref{icra:tab:subject-generalization} \& \ref{icra:tab:clutter-generalization}, we introduce
unseen clutter that interferes with the radar signal, including a metal box filled with daily
objects and a large paper box that blocks the sensor's line of sight. The success ratios are only
slightly affected in both cases. We also evaluate two unseen subjects with different heights, body
shapes, and waving habits to verify that the perception model does not overfit the participants in
the training dataset; performance remains stable across these subject changes. 



\section{Conclusion}
\label{icra:sec:conclusion}

This work presents mmHRI, a privacy-preserving human--robot interaction framework that integrates
mmWave radar-based human perception with a vision--language--action policy. By combining
motion-sensitive raw radar representations with the geometric information in radar point clouds,
mmHRI jointly estimates human actions, 3D poses and locations, and converts the resulting
human state into structured instructions for closed-loop robot control. The proposed
DRP and MSSM mechanism further improve robustness to curtain motion and robot-arm interference. Experiments show that mmHRI achieves reliable action recognition: 85.09\% accuracy
under occlusion and 74.91\% accuracy cross-environment. Real-world trials further demonstrate robust object
delivery and retrieval across unseen subjects, clutter configurations, and environments.
These results demonstrate the potential of mmWave radar as a complementary sensing modality for
privacy-sensitive HRI.

\noindent\textbf{Limitations.}
(1) The current system does not consider physical interaction with the curtain, such as opening curtains
before delivery. Instead, it focuses on perceiving human intent through the curtain and satisfying human requests. Such interaction could be extended with more demo data. (2) The current system still cannot grasp objects using mmWave radar alone or handle multiple subjects. Future work could explore radar-only HRI or multi-subject human perception.

\bibliographystyle{IEEEtran}
\bibliography{references}
\end{document}